\documentclass[11pt]{article}

\usepackage[preprint]{acl}

\usepackage{times}
\usepackage{latexsym}
\usepackage[T1]{fontenc}
\usepackage[utf8]{inputenc}
\usepackage{microtype}
\usepackage{inconsolata}

\usepackage{amsmath}
\usepackage{amssymb}
\usepackage{booktabs}
\usepackage{graphicx}
\usepackage{multirow}
\usepackage{xcolor}
\usepackage{placeins}
\usepackage{stfloats}

\newcommand{\navbind}{\textsc{NavBind}}
\newcommand{\symkl}{\ensuremath{D_{\mathrm{SKL}}}}

\newcommand{\ppl}{\ensuremath{\mathrm{PPL}}}
\newcommand{\attgrad}{\ensuremath{\left|A \odot \nabla_A Q_{h,s}\right|}}

\newcommand{\tblhead}[1]{\multicolumn{1}{c}{#1}}
\newcommand{\tblheadlast}[1]{\multicolumn{1}{c@{}}{#1}}

\title{\navbind: Diagnosing Instruction-Action Binding at Route Forks in Navigation VLAs}

\author{
  \textbf{Josh Qixuan Sun\textsuperscript{1}},
  \textbf{Xiaoying Xing\textsuperscript{2}},
  \textbf{Yurun Chen\textsuperscript{3,4}}, \\
  \textbf{Huaiyuan Weng\textsuperscript{1}},
  \textbf{Wentao Zhu\textsuperscript{4}},
  \textbf{Mark Crowley\textsuperscript{1}} \\
  \\
  \textsuperscript{1}University of Waterloo \\
  \textsuperscript{2}Northwest University \\
  \textsuperscript{3}Shanghai Jiao Tong University \\
  \textsuperscript{4}Eastern Institute of Technology, Ningbo \\
}

\begin{document}
\maketitle

\begin{abstract}
Vision-language-action (VLA) navigation models translate visual observations and instructions into actions.
Navigation success assesses whether an agent reaches the goal.
However, reaching the goal successfully does not establish whether the model genuinely follows the instructions or relies on visual cues or navigation history.
We introduce \navbind{} to diagnose instruction-action binding: whether changing the instruction influences the probabilistic distribution of the actions, with the navigation context fixed.
We construct 3,252 route-fork contexts by recombining route-aligned instruction segments from R2R-CE and RxR-CE.
Each context pairs full instructions that share the description of the route already traveled but specify different branches.
Both are evaluated under the same agent state and navigation history.
We compare generated local action sequences, the likelihoods of fixed reference action sequences under both instructions, and changes in visual attention and attribution.
Across the evaluated models, higher reported navigation success does not imply that the model more faithfully follows the instructions.
\end{abstract}

\section{Introduction}
\label{sec:introduction}

\begin{figure*}[t]
    \centering
    \includegraphics[width=\textwidth]{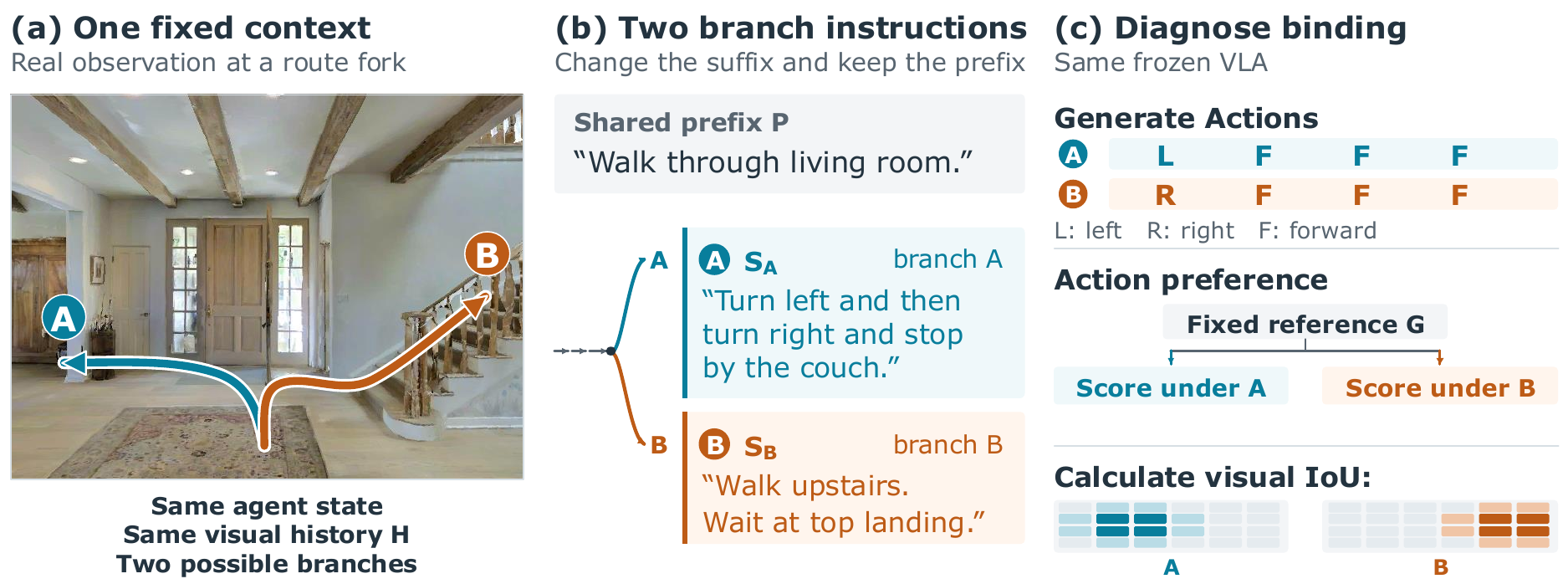}
    \caption{\textbf{Does the action follow the instruction?}
    \navbind{} pairs one fixed navigation history $H$ with two full instructions
    $P+S_A$ and $P+S_B$. The shared prefix $P$ describes the completed route,
    and the remaining suffixes $S_A$ and $S_B$ specify different executable branches.
    This example contrasts turning left with going up the stairs.
    We jointly examine generated action-sequence changes, fixed-target action
    likelihood, visual attention, and attribution for each instruction's
    corresponding reference actions.}
    \label{fig:overview}
\end{figure*}

In vision-and-language navigation in continuous environments (VLN-CE) \citep{krantz2020beyond}, vision-language-action (VLA) models \citep{zhang2024navid} map visual observations and instructions to actions.
Success rate measures whether an agent reaches the goal but does not establish whether it follows the instruction.
Route-fidelity metrics \citep{jain-etal-2019-stay} assess the similarity between the generated trajectory and the ground-truth trajectory, providing an indirect measure of instruction following.
However, these metrics still cannot establish \emph{whether the model genuinely follows the instructions}, as it may instead rely on visual cues or navigation history alone.

Language interventions, i.e., changing the language instruction and observing how the local actions or navigation performance change, provide a more direct test of language dependence.
NavTrust \citep{jiang2026navtrust} and Contrast Sets \citep{anwar2025contrast} rephrase instructions and assess changes in navigation performance.
However, these tests keep the intended navigation behaviour unchanged.
R2RIE-CE \citep{taioli2024mind} modifies instructions by injecting direction, object, and room errors.
However, it neither ensures that the modified instructions describe feasible routes nor provides reference actions corresponding to them.
Without matching reference trajectories for the modified instructions, a behavioural change alone does not show that the model followed them.
We therefore ask not merely whether the model's behaviour changes, but \emph{whether changing the instruction to describe another feasible route makes the model more likely to choose the corresponding actions}.
We keep the agent's state, navigation history, and completed instruction prefix fixed.
In Figure~\ref{fig:overview}, both turning left and continuing up the stairs are feasible, and the remaining instruction specifies which branch to take.
We use \emph{instruction-action binding} to describe the requirement that the model assigns higher probability to the actions required by the instruction.

We introduce \navbind, a controlled diagnostic built through cross-route
recombination of route-instruction examples drawn from R2R-CE and RxR-CE.
Using fine-grained route-language alignments, we retain one example's
completed instruction prefix and combine it with either its original suffix
or a compatible suffix from another route.
The resulting full-instruction pairs specify different executable
branches under the same replayed history and agent state, yielding
3,252 contexts with reference action sequences for both branches.

Within each fixed context, we first ask: does changing the instruction
change the action sequence the model generates?
We then ask: does the model assign higher probability to the actions
for a branch when instructed to take that branch rather than the other?
We test this by comparing the likelihood of each reference action
sequence under the two instructions.
This comparison can reveal changes in action preference even when the
generated sequence stays the same.
Finally, we examine the visual response: do attention heatmaps and
gradient-based attribution heatmaps highlight different image regions
under the two instructions?

We evaluate four navigation VLAs on \navbind{}: Uni-NaVid
\citep{zhang2025uninavid}, StreamVLN \citep{wei2025streamvln}, JanusVLN
\citep{zeng2026janusvln}, and FutureNav \citep{zhang2026futurenav}.
We find that higher reported navigation success does not imply that the model more faithfully follows the instructions.
(Table~\ref{tab:main-results}).
It also shows that models respond more strongly when the instructions
contrast opposite turns than when they contrast forward movement with a turn.

Our contributions are:
\begin{itemize}
    \item We construct \navbind{} by recombining instruction segments
    across routes into full instructions for two feasible branches.
    Each pair preserves the agent state, history, and
    completed instruction prefix, and provides reference action
    sequences for both branches.

    \item We diagnose instruction-action binding using generated
    sequences to measure output changes, fixed-target likelihoods
    to assess preferences for the instructed actions, and attention
    and gradient-based attribution heatmaps to track spatial changes.

    \item We show that higher reported navigation success does not imply that the model more faithfully follows the instructions. This supports evaluating how
    language affects local decisions alongside navigation success.
\end{itemize}

\section{Related Work}
\label{sec:related-work}

\subsection{Language Diagnostics in VLN}

\paragraph{Instruction fidelity beyond success.}
Goal completion can obscure whether an agent followed the described route.
Stay on the Path introduced CLS and longer R4R routes
\citep{jain-etal-2019-stay}, while LAW extended language-aligned supervision
and evaluation to VLN-CE \citep{raychaudhuri-etal-2021-language}.
Modality ablations further showed that R2R agents could rely strongly on graph
structure \citep{hu-etal-2019-looking}.  These works motivate
language-specific evaluation through task metrics, supervision, and modality
ablations.

\paragraph{Language interventions in VLN.}
Early diagnostics mask object, direction, and numerical cues
\citep{zhu-etal-2022-diagnosing} and swap directional tokens
\citep{hahn2023which}.
Skill-focused evaluations append atomic instructions to truncated RxR
episodes \citep{yang2023behavioral} and construct human-refined atomic
instructions, including same-start left/right pairs
\citep{wang-etal-2024-navigating}.
These studies evaluate agents on discrete navigation graphs and
their action tests cannot be directly applied to navigation VLAs
that execute low-level actions.

In VLN-CE, NavTrust evaluates whether agents maintain navigation performance
under instruction corruption and stylistic variation
\citep{jiang2026navtrust}.
Contrast Sets tests instruction rephrasing with unchanged navigation
goals on a physical robot
\citep{anwar2025contrast}.
However, these tests keep the intended navigation behaviour unchanged.
R2RIE-CE injects direction, object, and room errors into
instructions to study error detection and localization
\citep{taioli2024mind}.
The modified instructions are not paired with matching reference
trajectories.
However, it neither ensures that the modified instructions describe feasible routes nor provides reference actions corresponding to them.
Without matching reference trajectories for the modified instructions, a behavioural change alone does not show that the model followed them.
CAST pairs generated alternative instructions with short action
branches for counterfactual training \citep{glossop2026cast}, but the intended action behaviour only include a short horizon rather than a long horizon.

\navbind{} constructs full-instruction pairs for long-horizon navigation
by recombining original route-aligned instruction spans across source
routes.
The pairs specify different executable branches under the same
agent state, navigation history, and completed instruction prefix.
Scoring both reference action sequences under both instructions tests
whether the model's local action preferences shift toward the
instructed branch.

\subsection{Language-Action Coupling in VLAs}

Recent work examines whether language and other multimodal signals influence
VLA action generation.  VLA-Trace connects representation similarity,
attention-path knockouts, and rollout-level behavioural probes
\citep{shi2026vlatrace}, while \citet{haon2025mechanistic} identify semantic
activation directions and intervene on them to steer actions.
\citet{zhang2026igar} diagnose linguistic blindness in manipulation VLAs using
contradictory instructions and recalibrate attention at inference time, while
\citet{zhan-etal-2026-stable} study robustness to linguistic perturbations and
strengthen language guidance through residual steering.  These studies
establish that language and internal pathways can affect VLA actions, but their
evaluations center on manipulation and often combine diagnosis with model
modification.  NavBind tests instruction-action binding in frozen navigation
VLAs: two valid route instructions share the same history but specify different
executable branches.  We align behaviour with fixed-target likelihood and
visual attribution.

\section{Problem Setting}
\label{sec:problem-setting}

\subsection{VLN-CE}

In vision-and-language navigation in continuous environments (VLN-CE), an
agent follows a natural-language instruction $I$ through a continuous 3D
environment \citep{krantz2020beyond}.
At time $t$, it receives an egocentric observation $o_t$ and acts from the
navigation history
\begin{equation}
    H_t=(o_0,a_0,\ldots,a_{t-1},o_t).
\end{equation}
A navigation policy assigns a conditional distribution
$\pi_\theta(a_t\mid I,H_t)$ over its next one or few actions.  A full episode ends when the
agent emits \textsc{Stop} or reaches the maximum steps, and success means stopping near the destination.

We study the widely used four-action interface
\begin{equation}
    \mathcal{A}=\{\textsc{Forward},\textsc{Left},
    \textsc{Right},\textsc{Stop}\},
\end{equation}
adopted by low-level VLN-CE agents \cite{krantz2020beyond} and recent navigation VLAs
\citep{zhang2025uninavid,wei2025streamvln,zeng2026janusvln,zuo2026fantasyvln,yang2026gavln,zhang2026futurenav}.
\textsc{Forward} advances the agent by a fixed distance, while
\textsc{Left} and \textsc{Right} rotate it by a fixed angular increment.
This finite action vocabulary provides a common behavioural interface and, for
models that expose action logits, permits exact likelihood evaluation of a
fixed action sequence.

\subsection{Local Route-Fork Diagnosis}
\label{sec:local-fork}

\begin{figure}[t]
    \centering
    \includegraphics[width=\columnwidth]{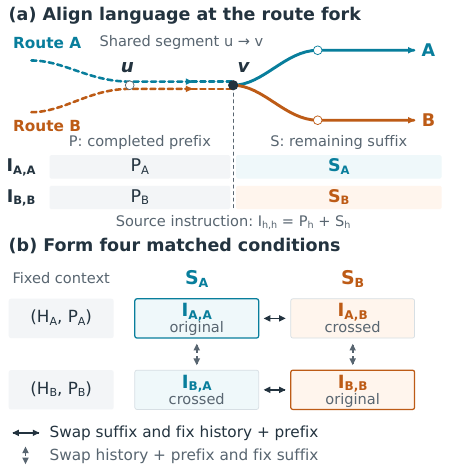}
    \caption{\textbf{Route-aligned pair construction.}
    (a) Original routes A/B (blue/orange) share a directed segment from $u$ to
    fork $v$, then diverge.
    Dashed paths are traversed, and solid paths remain. Each full instruction
    $I_{h,h}=P_h+S_h$ splits at $v$ into completed prefix $P_h$ and remaining suffix $S_h$.
    (b) Horizontal comparisons change only the suffix.
    Vertical comparisons change history and prefix together with the suffix
    fixed. }
    \label{fig:construction}
\end{figure}

Within each source pair, a local diagnostic context consists of a valid
history $H_h$ ending at a nonterminal route fork $v$ and its completed
instruction prefix $P_h$, where $h\in\{A,B\}$ identifies the source route
(Figure~\ref{fig:construction}(a)). Each remaining suffix $S_s$, with
$s\in\{A,B\}$, specifies one outgoing branch. The full input instruction is
$I_{h,s}=P_h+S_s$, and an evaluation condition consists of $H_h$ and $I_{h,s}$.

From the state reached by $H_h$, the reference action sequence for either
executable branch $g\in\{A,B\}$ is
\begin{equation}
    G_{h,g}=(a^{h,g}_1,\ldots,a^{h,g}_{K_{h,g}}),\quad K_{h,g}\leq K.
\end{equation}
The diagnostic horizon $K$ limits its length, and \textsc{Stop} ends the
sequence.

For each context $(H_h,P_h)$, \navbind{} compares $I_{h,A}$ and $I_{h,B}$,
as in a horizontal comparison in Figure~\ref{fig:construction}(b).
Both conditions start from exactly the same pose, current observation, and
observation-action history.
Instruction-action binding requires the model to assign higher probability
to the actions required by the suffix. Section~\ref{sec:construction}
describes how route geometry and fine-grained language annotations yield
these matched conditions.

\section{Counterfactual Language Pairs for Navigation}
\label{sec:construction}

\subsection{Route-Aligned Source Pairs}

We construct source pairs from English R2R-CE and RxR-CE routes and instructions
\citep{anderson2018vln,ku2020rxr,krantz2020beyond}.
For R2R and English RxR, we use released fine-grained annotations from
FGR2R \citep{hong2020sub} and Landmark-RxR \citep{he2021landmark},
respectively. These annotations divide full instructions into shorter
text segments, called \emph{sub-instructions}, and align each segment
with the part of the trajectory it describes.

We first identify route pairs with the structure shown in
Figure~\ref{fig:construction}(a): two routes in the same scene share a directed
segment from $u$ to the fork $v$, then take different outgoing edges at $v$.
The routes may differ before $u$. Both branches beyond $v$ must be executable
and nonterminal.  We require a sufficiently long shared segment $u\!\to\!v$
to align the two routes' positions and orientations
at $v$.  Sharing $v$ alone leaves arrival
orientation unconstrained, which can change the interpretation of
instructions such as ``turn left.''

For each source route $h$ in a candidate pair, we use the sub-instruction
annotations to place the language boundary at $v$. An edge-exact boundary
requires the preceding subpath to end at $v$ and the next subpath to start at $v$ with its
first edge entering branch $h$. We recover the corresponding
original-text offsets by monotonic matching with case, spacing, and punctuation
normalized.  Cutting at the start of that next sub-instruction yields
$I_{h,h}=P_h+S_h$: the prefix $P_h$ covers the full route up to $v$, including
any portion before $u$, and the suffix $S_h$ retains the remaining instruction.
Both spans preserve their original wording.
Selection uses route annotations alone. Repeated physical forks are deduplicated.

For each retained pair, expert replay of route $h$ from its source start to $v$
yields a history $H_h$ paired with its completed prefix $P_h$.  From each state at $v$,
we construct both local branch targets $G_{h,A}$ and $G_{h,B}$.
The resulting history-prefix pairs form the fixed
row contexts in Figure~\ref{fig:construction}(b).

\subsection{Matched Interventions}

Figure~\ref{fig:construction}(b) crosses each context $(H_h,P_h)$ with both
suffixes. Each cell gives the instruction $I_{h,s}$, evaluated with the
history $H_h$ in its row.

\paragraph{Suffix intervention.}
Within each row, the diagonal instruction $I_{h,h}$ retains the source text,
and the off-diagonal instruction substitutes the other route's suffix.
Evaluating $I_{h,A}$ and $I_{h,B}$ from the same replay history and fork state
implements the primary fixed-context
comparison defined in Section~\ref{sec:local-fork}.

\paragraph{Matched prefix intervention.}
Across rows, the supplementary comparison between $(H_A,I_{A,s})$ and
$(H_B,I_{B,s})$ holds the suffix fixed while jointly changing the history
and its source prefix.
It compares the model's response to the same branch beyond $v$ across
the two source-matched contexts.

Each source pair initially yields two candidate contexts and four conditions,
linked by a source-pair identifier. We retain contexts whose two reference
sequences differ within four actions at every evaluated native action scale.
The resulting \navbind{} set contains 3,252 contexts from 1,648 source pairs,
shared by all models. 

\section{Diagnosing Instruction-Action Binding}
\label{sec:diagnostics}

\begin{figure}[!t]
    \centering
    \includegraphics[width=\columnwidth]{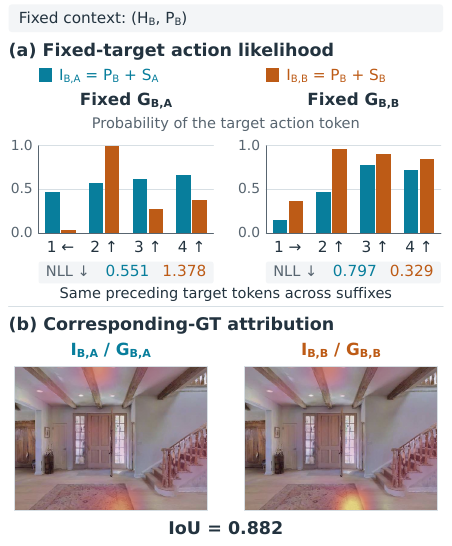}
    \caption{\textbf{Action likelihood and visual attribution.}
    (a) Conditional action-token probabilities for each fixed reference
    under the two suffixes.
    Arrows denote left, forward, and right actions.
    (b) Full attention$\times$gradient heatmaps for each suffix's
    corresponding reference. Warmer colors indicate higher attribution
    on a shared scale. IoU uses the original top-25\% masks.}
    \label{fig:diagnostics}
\end{figure}

Given the matched conditions constructed in Section~\ref{sec:construction}, we
analyze instruction-action binding from three complementary perspectives.  We
first measure whether an intervention changes the generated local action
sequence. We then measure how it changes the likelihood and
action distribution of a fixed reference target. Finally, we examine
how visual attention and attribution change across conditions. Likelihood
comparisons keep the target fixed, while gradient-based attribution follows
each instruction's corresponding GT. Figure~\ref{fig:diagnostics} illustrates
these comparisons on one navigation context.

\subsection{Generated Action Sequences}

For each context $(H_h,P_h)$, we generate local action sequences under
$I_{h,A}$ and $I_{h,B}$ using native inference without teacher forcing.
We report the percentage of contexts whose sequences match exactly in action
symbols, order, and length within horizon $K$ as Same sequence
(Same seq.).

\subsection{Fixed-Target Action Likelihood}

Discrete generation can remain unchanged even when model's
action distribution changes. We therefore measure
changes in the likelihood of fixed reference action sequences.

Figure~\ref{fig:diagnostics}(a) illustrates the scoring procedure for context
$(H_B,P_B)$, corresponding to the second row of Figure~\ref{fig:construction}(b).
The targets $G_{B,A}$ and $G_{B,B}$ are each scored under both instructions,
$I_{B,A}$ and $I_{B,B}$, giving four likelihood scores.

In general, $L_{h,s,g}$ denotes the mean negative log-likelihood (NLL) of
target $G_{h,g}$ under instruction $I_{h,s}$ and history $H_h$. Its indices
specify the history $h$, input suffix $s$, and target branch $g$, respectively.
For chunk policies, we compute the mean NLL over the target's action tokens
$z_1,\ldots,z_N$:
\begin{equation}
    L_{h,s,g}=-\frac{1}{N}\sum_{k=1}^{N}
      \log p_\theta\!\left(z_k\mid I_{h,s},H_h,z_{<k}\right).
\end{equation}
Here $p_\theta$ is the model's full-vocabulary token probability. We use
natural logarithms throughout. At position
$k$, the model receives the same preceding target tokens $z_{<k}$ under both suffixes
(teacher forcing). Only action tokens contribute to the average.
Lower $L_{h,s,g}$ means higher target likelihood, and perplexity is
$\ppl_{h,s,g}=\exp(L_{h,s,g})$.

Single-step policies instead score each target action at the state reached
by executing its preceding GT actions. We sum the native action-token log
probabilities across decisions and divide by the total valid token count.
Both suffixes use the same GT observation sequence, with independent model
memory. This supplies a closed-loop counterpart to the chunk score. The primary suffix effects keep $H_h$, $P_h$, and $G_{h,g}$ fixed:
\begin{align}
    d_{h,A}&=L_{h,B,A}-L_{h,A,A}, \\
    d_{h,B}&=L_{h,A,B}-L_{h,B,B}.
\end{align}
The mean is
$\bar d_h=\tfrac{1}{2}(d_{h,A}+d_{h,B})$. We report its average across
evaluated samples as $\bar d$.

To interpret $\bar d$ in Table~\ref{tab:main-results}, positive values mean
that matching suffixes make their reference actions more likely,
negative values mean less likely, and zero means no net advantage.
We convert this log-probability difference to a relative percentage change
using $(\exp(\bar d)-1)\times100\%$. Thus, $\bar d=0.0127$ gives
$(\exp(0.0127)-1)\times100\%\approx1.28\%$ for reference-action token
probabilities, averaged geometrically across tokens, targets, and contexts.

We report Both+ as the percentage of contexts where both $d_{h,A}>0$ and
$d_{h,B}>0$, so each reference is more likely under its matching suffix.

As a complementary metric, we report symmetric KL divergence. Let
$q^{(j)}_{h,s,g}$ denote the action distribution, renormalized over the four
legal actions, at reference step
$j$ of $G_{h,g}$, scored with $I_{h,s}$ and $H_h$ using the protocol above.
With the same preceding target actions, we compute
$\symkl(q^{(j)}_{h,A,g},q^{(j)}_{h,B,g})$, then average over steps, targets,
and contexts. KL is zero for unchanged action probabilities and increases
with their redistribution, irrespective of target alignment.

\subsection{Visual Attention and GT-Conditioned Attribution}

We next compare visual attention and the visual sensitivity of each
instruction's corresponding reference action sequence. Let $A$ denote decoder
attention from action-query positions to visual tokens. Raw attention is
measured under each fixed reference sequence $G_{h,g}$, so both suffixes
receive the same preceding target tokens at every prediction position.

For gradient-based attribution, the explained action follows the instruction:
$I_{h,s}$ is paired with its corresponding GT sequence $G_{h,s}$. Its score is
\begin{equation}
    Q_{h,s}=-L_{h,s,s}.
\end{equation}
We extract gradient magnitude $|\nabla_A Q_{h,s}|$ and the elementwise
product magnitude $\attgrad$. For chunk policies, maps aggregate the
teacher-forced action-query positions under fixed observations.

Let $M^{e}_{h,s,g}$ denote a map computed under instruction $I_{h,s}$ and
reference sequence $G_{h,g}$. Raw attention compares the same reference
under both suffixes:
\begin{equation}
    \operatorname{IoU}^{\mathrm{attn}}_{h,g}=
    \operatorname{IoU}\!\left(
      \mathcal{B}(M^{\mathrm{attn}}_{h,A,g}),
      \mathcal{B}(M^{\mathrm{attn}}_{h,B,g})
    \right),
\end{equation}
and averages the two references within each context. Gradient and
attention$\times$gradient compare the corresponding GTs:
\begin{equation}
    \operatorname{IoU}^{e}_{h}=\operatorname{IoU}\!\left(
      \mathcal{B}(M^{e}_{h,A,A}),\mathcal{B}(M^{e}_{h,B,B})
    \right).
\end{equation}
This comparison jointly changes the instruction and the explained action.
Here $\mathcal{B}$ selects a fixed top-response fraction on the scoring grid.
Heatmaps are interpolated to RGB
resolution for display, while IoU uses the original scoring-grid masks.

For single-step policies, fixed-reference attention is compared at aligned
GT states and averaged over steps. Corresponding-GT gradients are compared
only at the shared fork observation, using each instruction's first GT
action. Later GT trajectories can have different views.

IoU equals one for identical selected regions and zero for disjoint regions.
Lower IoU indicates greater spatial change under the respective comparison.
Section~\ref{sec:results} relates these changes to fixed-target likelihoods
and changes in the action sequences.

\section{Experimental Setup}
\label{sec:experimental-setup}

\begin{table*}[t]
\centering
\small
\setlength{\tabcolsep}{4pt}
\begin{tabular}{lrrrrrrrr}
\toprule
& \tblhead{RxR-CE} & \tblhead{Behavior}
& \multicolumn{3}{c}{Fixed-target likelihood}
& \multicolumn{3}{c}{Visual overlap (IoU)} \\
\cmidrule(lr){2-2}\cmidrule(lr){3-3}\cmidrule(lr){4-6}\cmidrule(lr){7-9}
Model & \tblhead{\shortstack{SR\\(\%) $\uparrow$}}
& \tblhead{\shortstack{Same seq.\\(\%) $\downarrow$}}
& \tblhead{$\bar d\uparrow$}
& \tblhead{\shortstack{Both+\\(\%) $\uparrow$}} & \tblhead{KL $\uparrow$}
& \tblhead{Attn. $\downarrow$} & \tblhead{Grad. $\downarrow$} & \tblhead{Attn.$\times$Grad. $\downarrow$} \\
\midrule
Uni-NaVid & 56.1 & 85.79 & 0.0127 & 29.15 & 0.0124 & 0.860 & 0.454 & 0.773 \\
StreamVLN & 54.4 & 32.41 & 0.8360 & 80.38 & 0.8233 & 0.602 & 0.420 & 0.456 \\
JanusVLN$^{*}$ & 51.4 & 30.75 & 0.3643 & 72.88 & 0.4537 & 0.727 & 0.760 & 0.698 \\
FutureNav$^{*}$ & 54.5 & 25.49 & 0.4411 & 70.60 & 0.6208 & 0.728 & 0.786 & 0.702 \\
\bottomrule
\end{tabular}
\caption{Reported navigation success and \navbind{} diagnostics.
SR is the officially reported RxR-CE val-unseen success rate for each
released checkpoint used in our experiments. Lower Same seq. and IoU indicate greater output
and spatial change. Higher $\bar d$ and Both+ indicate
stronger or more frequent matching-suffix likelihood gains.
Higher KL indicates greater action-distribution change.
IoU uses top-25\% masks. $^{*}$ marks single-step models.}
\label{tab:main-results}
\end{table*}

\subsection{Data and Models}

We evaluate Uni-NaVid \citep{zhang2025uninavid}, StreamVLN
\citep{wei2025streamvln}, JanusVLN \citep{zeng2026janusvln}, and FutureNav
\citep{zhang2026futurenav} on the same 3,252 \navbind{} contexts constructed in
Section~\ref{sec:construction}, using released checkpoints.

\subsection{Evaluation Protocol}

We use a four-action horizon ($K=4$) and each model's native action
increments. Uni-NaVid and StreamVLN generate action chunks under fixed
observations, while JanusVLN and FutureNav execute each predicted action
before the next decision. We retain native decoding: Uni-NaVid uses sampling
and StreamVLN uses greedy decoding. 

\section{Results}
\label{sec:results}
\label{sec:analysis}

\subsection{Main Results}

In Table~\ref{tab:main-results}, Uni-NaVid has the highest reported
RxR-CE val-unseen SR. However, StreamVLN has the highest $\bar d$ and
KL values.
Uni-NaVid's Both+ is 29.15\%, only 4.15 percentage points above a 25\%
reference. This reference assumes that the two likelihood gains are
independent and each has a 50\% chance of being positive.
StreamVLN and FutureNav have nearly identical reported SR (54.4\%
and 54.5\%), yet StreamVLN has higher $\bar d$ (0.8360 versus 0.4411),
Both+ (80.38\% versus 70.60\%), and KL (0.8233 versus 0.6208).
Thus, higher navigation success does not necessarily mean that changing
the instruction has a larger effect on action probabilities.
JanusVLN and FutureNav change their action sequences more often than
StreamVLN (lower Same seq.), but show smaller likelihood gains for the
instructed action sequences (lower $\bar d$). Their attention and attribution
heatmaps also highlight more similar visual regions across instructions
(higher IoU).
One possible explanation is that these models rely on similar visual
cues without clearly distinguishing the instructed branch, leaving them
less confident about which action to take.
Small probability shifts could then be enough to change the selected action.

\subsection{Consistency Across Diagnostics}

Table~\ref{tab:joint-results} shows that StreamVLN's Unchanged group
has lower $\bar d$ than its Changed group.
In this group, changing the instruction produces a smaller increase
in the likelihood of the required actions.
The Unchanged group also has higher Attn., Grad., and Attn.$\times$Grad.
IoU, so the heatmaps highlight more similar regions under the two
instructions.
Thus, the likelihood and visual results agree with the generated-action
results: unchanged action sequences come with smaller changes toward
the required actions and smaller changes in the highlighted visual regions.

\begin{table}[!htbp]
\centering
\small
\setlength{\tabcolsep}{2pt}
\begin{tabular}{lrrrrr}
\toprule
Output & \tblhead{N} & \tblhead{$\bar d\uparrow$}
& \tblhead{\shortstack{Attn.\\IoU $\downarrow$}}
& \tblhead{\shortstack{Grad.\\IoU $\downarrow$}}
& \tblhead{\shortstack{Attn.$\times$Grad.\\IoU $\downarrow$}} \\
\midrule
Unchanged & 1,054 & 0.3766 & 0.6754 & 0.4831 & 0.5128 \\
Changed & 2,198 & 1.0563 & 0.5673 & 0.3895 & 0.4292 \\
\bottomrule
\end{tabular}
\caption{StreamVLN results for unchanged and changed action sequences
under the two instructions.}
\label{tab:joint-results}
\end{table}

\FloatBarrier
\subsection{Weaker Responses with Longer Prefixes}
\label{sec:prefix-tokens}

Table~\ref{tab:prefix-tokens} shows that StreamVLN has higher Same seq.
and lower $\bar d$ and KL as the completed prefix contains more tokens.
With longer completed prefixes, the model more often generates the same
action sequence under the two instructions, and the overall action distribution changes less.
Thus, longer completed prefixes are associated with weaker responses
to the remaining instruction.

\begin{table}[!htbp]
\centering
\small
\setlength{\tabcolsep}{4pt}
\begin{tabular}{@{}lrrr@{}}
\toprule
Prefix tokens & \tblhead{\shortstack{Same seq.\\(\%) $\downarrow$}}
& \tblhead{$\bar d\uparrow$} & \tblheadlast{KL $\uparrow$} \\
\midrule
$\leq 47$ & 30.80 & 0.8904 & 0.8964 \\
$48$-$76$ & 30.90 & 0.8492 & 0.8497 \\
$\geq 77$ & 35.62 & 0.7660 & 0.7205 \\
\bottomrule
\end{tabular}
\caption{StreamVLN results by prefix token count, with roughly equal group sizes.}
\label{tab:prefix-tokens}
\end{table}

\FloatBarrier
\subsection{Stronger Responses to Opposite Turns}
\label{sec:action-contrasts}

We compare StreamVLN across three action-pair groups: forward-left (F/L),
forward-right (F/R), and left-right (L/R).
Table~\ref{tab:streamvln-action-contrasts} shows that the L/R group has lower
Same seq. and higher $\bar d$ and KL than the other two groups.
In the L/R group, changing the instruction changes the generated action
sequence more often. It also produces a
larger increase in the likelihood of the required actions and a larger
change in the overall action distribution.
Thus, StreamVLN responds more strongly when the instructions
contrast opposite turns than when they contrast forward movement with a turn.

\begin{table}[!htbp]
\centering
\small
\setlength{\tabcolsep}{5pt}
\begin{tabular}{@{}lrrrr@{}}
\toprule
Action pair & \tblhead{$N$} & \tblhead{\shortstack{Same seq.\\(\%) $\downarrow$}}
& \tblhead{$\bar d\uparrow$} & \tblheadlast{KL $\uparrow$} \\
\midrule
F/L & 486 & 37.24 & 0.7426 & 0.6905 \\
F/R & 508 & 33.07 & 0.8404 & 0.8248 \\
L/R & 1,504 & 26.66 & 1.0135 & 0.9458 \\
\bottomrule
\end{tabular}
\caption{StreamVLN results by action-pair group.
F/L/R denote forward/left/right.}
\label{tab:streamvln-action-contrasts}
\end{table}

\FloatBarrier
\subsection{Attention Changes at Action Divergence}
\label{sec:action-position-attention}

We group contexts by the first step at which the two GT sequences differ.
For example, the GT sequences $[F,F,L,F]$ and $[F,F,R,F]$ first differ at
step 3. Table~\ref{tab:streamvln-action-position-attention} reports
StreamVLN's attention IoU at each of the four action positions.
For contexts whose GT sequences first differ at step 2, 3, or 4, attention
IoU is higher at the preceding positions and reaches its lowest value at
the first differing position. In the step-2 and step-3 groups, IoU rises
again at the next position.
Thus, after shared initial actions, the attention
heatmaps under the two instructions differ most where the required actions
first differ.

\begin{table}[!htbp]
\centering
\small
\setlength{\tabcolsep}{3pt}
\begin{tabular}{@{}lrrrrr@{}}
\toprule
\multirow{2}{*}{\shortstack[l]{First differing\\GT position}}
& \tblhead{\multirow{2}{*}{$N$}} & \multicolumn{4}{c@{}}{Action position} \\
\cmidrule(lr){3-6}
& & \tblhead{1} & \tblhead{2} & \tblhead{3} & \tblheadlast{4} \\
\midrule
1 & 2,498 & 0.5911 & 0.5829 & \textbf{0.5687} & 0.5757 \\
2 & 409 & 0.6103 & \textbf{0.5693} & 0.5736 & 0.5765 \\
3 & 219 & 0.6569 & 0.6183 & \textbf{0.5777} & 0.6021 \\
4 & 126 & 0.6792 & 0.6642 & 0.6373 & \textbf{0.6273} \\
\bottomrule
\end{tabular}
\caption{StreamVLN attention IoU at each action position.
Bold marks the lowest mean IoU in each row.}
\label{tab:streamvln-action-position-attention}
\end{table}

\FloatBarrier
\subsection{Larger Original-Pairing Advantages on Training Sources}
\label{sec:original-pairing}

StreamVLN has lower NLL under original instructions than under recombined
instructions, with a larger gap on train than on val-unseen.
For each context $(H_h,P_h)$, we compare $L_{h,h,h}$ under the original
instruction with $L_{h,s,s}$ under the recombined instruction ($s\ne h$),
scoring each instruction's corresponding reference. We report results
separately for train and val-unseen, with both source instructions in each
context coming from the same split. The gap is mean recombined NLL minus mean original NLL.
A positive gap means lower NLL under original instructions.

\begin{table}[!htbp]
\centering
\small
\setlength{\tabcolsep}{3pt}
\begin{tabular}{@{}lrrrr@{}}
\toprule
Source split & \tblhead{$N$} & \tblhead{Original $\downarrow$}
& \tblhead{Recombined $\downarrow$} & \tblheadlast{Gap} \\
\midrule
Train & 2,367 & 0.4666 & 0.6007 & +0.1341 \\
Val-unseen & 385 & 0.5763 & 0.6374 & +0.0611 \\
\bottomrule
\end{tabular}
\caption{StreamVLN mean reference-action NLL under original and
recombined pairings. Gap is recombined minus original NLL.}
\label{tab:streamvln-original-recombined}
\end{table}

Table~\ref{tab:streamvln-original-recombined} shows lower NLL under original
instructions than under recombined instructions in both splits.
The gap is larger on train than on val-unseen.
StreamVLN has lower NLL on train than on val-unseen, but this difference
is smaller for recombined instructions than for original instructions.
This pattern is consistent with overfitting to the original pairings.
\FloatBarrier

\subsection{Stability Across Histories}
\label{sec:cross-history}

We compare the two source histories and their matched prefixes
(Figure~\ref{fig:construction}(b)). With the suffix and four-action reference
unchanged, we expect consistent generated actions and associated visual
evidence across histories. Here, KL compares the action distributions
under the two histories for the same reference.

\begin{table}[!htbp]
\centering
\small
\setlength{\tabcolsep}{1.7pt}
\begin{tabular}{@{}lrrrrr@{}}
\toprule
Model & \tblhead{\shortstack{Same seq.\\(\%) $\uparrow$}}
& \tblhead{KL $\downarrow$}
& \tblhead{\shortstack{Attn.\\IoU $\uparrow$}}
& \tblhead{\shortstack{Grad.\\IoU $\uparrow$}}
& \tblheadlast{\shortstack{Attn.$\times$Grad.\\IoU $\uparrow$}} \\
\midrule
StreamVLN & 51.42 & 0.3131 & 0.4295 & 0.3032 & 0.4088 \\
JanusVLN$^{*}$ & 25.18 & 0.3225 & 0.2675 & 0.2278 & 0.2543 \\
FutureNav$^{*}$ & 27.67 & 0.3150 & 0.3083 & 0.2548 & 0.2894 \\
\bottomrule
\end{tabular}
\caption{Cross-history stability. With the suffix and reference actions
fixed, higher Same seq. and IoU and lower KL indicate greater stability,
opposite to Table~\ref{tab:main-results}. $^{*}$ marks single-step models.}
\label{tab:cross-history}
\end{table}

Table~\ref{tab:cross-history} shows that StreamVLN preserves its generated
sequence more often than JanusVLN and FutureNav (51.42\% versus 25.18\%
and 27.67\%). It also has higher IoU across all three visual measures
and slightly lower KL.
Thus, among these three models, generated actions, action distributions,
and visual heatmaps give the same stability ordering: StreamVLN,
FutureNav, and JanusVLN.
\FloatBarrier

\section{Conclusion}
\label{sec:conclusion}

\navbind{} diagnoses instruction-action binding through route-fork
interventions under fixed navigation contexts, combining generated behavior,
fixed-target likelihoods, and visual attention and attribution. Across 3,252
shared contexts, higher reported navigation success does not always mean
more frequent action changes or larger shifts in action probabilities.
These findings support evaluating how language affects local decisions
alongside episode-level navigation success.

\section*{Limitations}
\label{sec:limitations}

Recombining source instructions may introduce ambiguity
or unnatural transitions at the segment boundaries. The four-action window
captures local instruction responses, while later responses and error recovery
remain outside the evaluation horizon. Model-native action increments also
give this window different physical extents across models. Visual IoU
quantifies spatial changes in attention and attribution, without establishing
whether the selected regions are relevant to the instructed branch or causally
determine the action.

\bibliography{refs}


\end{document}